\documentclass{article}

\PassOptionsToPackage{numbers,sort&compress}{natbib}

\usepackage[preprint]{neurips_2026}

\usepackage[utf8]{inputenc}
\usepackage[T1]{fontenc}
\usepackage{amsfonts}
\usepackage{amsmath}
\usepackage{newtxtext,newtxmath}
\usepackage{hyperref}
\usepackage{fontawesome5}
\usepackage{url}
\usepackage{booktabs}
\usepackage{nicefrac}
\usepackage{microtype}
\usepackage[table]{xcolor}
\usepackage{graphicx}
\usepackage{wrapfig}
\usepackage[font=small,labelfont=bf]{caption}
\usepackage{subcaption}
\usepackage{cancel}

\renewcommand{\arraystretch}{0.95}
\newcommand{\iacc}{\ensuremath{\mathrm{I}_{\mathrm{Acc}}}}
\newcommand{\acc}{\ensuremath{\mathrm{Acc}}}
\newcommand{\qacc}{\ensuremath{\mathrm{Q}_{\mathrm{Acc}}}}
\newcommand{\vacc}{\ensuremath{\mathrm{V}_{\mathrm{Acc}}}}

\newcommand{\TmainCaption}{%
  \caption{Open models fail; only a frontier model breaks away. Instance Accuracy
(\iacc; chance $6.25\%$) and per-item \acc\ across the three suites; Video-MME is
4-way multiple choice with no paired structure, so it reports \acc\ only (chance
$25\%$). On MotionBlind, \acc\ hovers near the $50\%$ coin-flip even when \iacc\ is
at the floor: models look competent item-by-item but cannot tell the paired clips
apart, while the same open models clear Video-MME comfortably ($58$--$70\%$ against
$25\%$ chance). Scores are each model's best combination of frame budget and sampling method per dataset. $^{*}$Evaluated on a representative $56$-instance subset ($224$ questions) of the $60$-instance suite; not $1{:}1$ comparable.\quad $^{\dagger}$reasoning model, evaluated with reasoning disabled on the $60$-instance set (provisional at $n{=}60$).\quad $^{\ddagger}$motion-tuned Qwen2.5-VL-7B.\quad $^{\P}$GPT-5.6 Luna at $720{\times}1280$, \texttt{detail=high}.\quad $^{\S}$MotionBlind human is the mean of $5$ annotators in our study ($60$ instances each; range $85$--$97$ \iacc); TimeBlind human from \citep{li2026timeblind}.\quad $^{\|}$The Gemini\,3.1\,Pro TimeBlind result is reported from Gemini\,3.0 in the original paper; Gemini\,3.0 is now deprecated, so we were unable to rerun the evaluation on the same model.\quad ``--'' = not computed for this work.}

  \label{tab:main}}

\newcommand{\TmainTab}{%
  \begin{tabular}{@{}l cc cc c@{}}
    \toprule
    \multicolumn{6}{@{}c@{}}{\emph{Best configuration per model across all frame sweeps}}\\
    \midrule
    & \multicolumn{2}{c}{\textbf{MotionBlind}} & \multicolumn{2}{c}{\textbf{TimeBlind}} & \textbf{Video-MME} \\
    \cmidrule(lr){2-3}\cmidrule(lr){4-5}\cmidrule(lr){6-6}
    \textbf{Model} & \iacc & \acc & \iacc & \acc & \acc \\
    \midrule
    \emph{Chance} & 6.25 & 50.0 & 6.25 & 50.0 & 25.0 \\
    \midrule
    \multicolumn{6}{@{}l}{\textit{Open Video-LLMs}}\\
    Eagle2.5-8B~\citep{chen2025eagle} & 11.7 & 58.8 & 24.2 & \textbf{66.5} & 65.4 \\
    Qwen3-VL-4B~\citep{qwen2025vl3} & 8.3 & 58.3 & \textbf{24.7} & 65.3 & 59.7 \\
    Motion-o (7B)~\citep{galoaa2026motiono}\,$^{\ddagger}$ & 6.7 & 55.4 & 16.3 & 61.0 & 58.2 \\
    Gemma-4-12B-it~\citep{gemmateam2026gemma4}\,$^{*}$ & 3.6 & 52.2 & 14.3 & 60.2 & 70.3 \\
    Molmo2-8B~\citep{ai2025molmo2,deitke2024molmo} & 3.3 & 54.2 & 12.2 & 59.0 & 60.9 \\
    Inkling-Small~\citep{inkling2026}\,$^{\dagger}$ & 10.0 & 57.1 & -- & -- & -- \\
    \midrule
    \multicolumn{6}{@{}l}{\textit{Frontier proprietary}}\\
    Gemini\,3.1\,Pro~\citep{gemini31pro2026} & \textbf{60.0} & \textbf{80.8} & 48.2\,$^{\|}$ & 76.2\,$^{\|}$ & \textbf{78.2} \\
    GPT-5.6 Luna~\citep{gpt56luna2026}\,$^{\P}$ & 15.0 & 62.1 & 46.3 & 77.3  & 74.8 \\
    \midrule 
    \emph{Human}\,$^{\S}$ & 91.3 & 97.8 & 98.2 & 99.3 & 87.9 \\
    \bottomrule
  \end{tabular}}

\newcommand{\TprobeCaption}{%
\caption{Integrity probes at a matched uniform-16 frame budget. We evaluate each benchmark on the original video, shuffled/reversed frames, and text-only input. On \textit{MotionBlind}, removing video or disrupting temporal order reduces all models to at or near $0\%$ \iacc, below the $6.25\%$ chance level, showing that success requires both visual evidence and correct temporal ordering. On \textit{TimeBlind}, disrupting order causes a $2$--$4\times$ drop, while models retain $2$--$7.5\%$ \iacc, indicating that some questions remain answerable without full temporal reasoning. In contrast, on \textit{Video-MME}, the same temporal order probes reduce performance by only $0.4$--$2.5$ accuracy points, with text-only performance remaining at approximately $39$--$44\%$, well above the $25\%$ chance level. Together, these results show that MotionBlind is uniquely dependent on temporal motion understanding with negligible language or prompt leakage.}
  \label{tab:probe}}

\newcommand{\TprobeTab}{%
  \begin{tabular}{@{}l *{4}{>{\centering\arraybackslash}p{2.5em}}@{}}
    \toprule
    \multicolumn{5}{@{}c@{}}{\emph{Matched budget: uniform, $N{=}16$}}\\
    \midrule
    \textbf{Model} & \textbf{Ord.} & \textbf{Shuf.} & \textbf{Rev.} & \textbf{\cancel{Vid.}} \\
    \midrule
    \multicolumn{5}{@{}l}{\textit{MotionBlind (\iacc)}}\\
    Eagle2.5-8B & 10.0 & 0.0 & 0.0 & 0.0 \\
    Qwen3-VL-4B & 6.7 & 0.0 & 0.0 & 0.0 \\
    Motion-o & 6.7 & 0.0 & 0.0 & 0.0 \\
    Molmo2-8B & 1.7 & 0.0 & 1.7 & 0.0 \\
    \midrule
    \multicolumn{5}{@{}l}{\textit{TimeBlind (\iacc)}}\\
    Eagle2.5-8B & 21.3 & 7.3 & 5.3 & 0.0 \\
    Qwen3-VL-4B & 23.7 & 7.5 & 4.5 & 0.0 \\
    Motion-o & 13.0 & 4.0 & 2.0 & 0.3 \\
    Molmo2-8B & 10.3 & 5.7 & 3.8 & 0.0 \\
    \midrule
    \multicolumn{5}{@{}l}{\textit{Video-MME (\acc; chance $25$)}}\\
    Eagle2.5-8B & 62.2 & 61.3 & 59.9 & 43.7 \\
    Qwen3-VL-4B & 56.6 & 54.9 & 54.6 & 41.6 \\
    Motion-o & 54.9 & 54.5 & 53.9 & 39.4 \\
    Molmo2-8B & 59.8 & 57.9 & 57.3 & 44.3 \\
    \bottomrule
  \end{tabular}}

\newif\ifcombinetables
\combinetablestrue          

\newcommand{\Tcombined}{%
\begin{table}[t]\centering
  \begin{minipage}[c]{0.57\linewidth}\footnotesize
    \setlength{\tabcolsep}{2.5pt}\renewcommand{\arraystretch}{1.12}%
    {\centering\TmainTab\par}
  \end{minipage}%
  \hfill
  \begin{minipage}[c]{0.41\linewidth}\footnotesize
    \setlength{\tabcolsep}{2.5pt}\renewcommand{\arraystretch}{1.12}%
    {\centering\TprobeTab\par}
  \end{minipage}
  \par\vspace{7pt}
  \begin{minipage}[t]{0.57\linewidth}\vspace{0pt}\footnotesize
    \TmainCaption
  \end{minipage}%
  \hfill
  \begin{minipage}[t]{0.41\linewidth}\vspace{0pt}\footnotesize
    \TprobeCaption
  \end{minipage}
  \par\vspace{6pt}
\end{table}}

\newcommand{\Tmain}{%
  \ifcombinetables \Tcombined \else
    \begin{table}[t]\centering\small
      \setlength{\tabcolsep}{6pt}\renewcommand{\arraystretch}{1.12}%
      {\centering\TmainTab\par}
      \vspace{5pt}
      \TmainCaption
      \par\vspace{4pt}
    \end{table}%
  \fi}

\newcommand{\Tprobe}{%
  \ifcombinetables\else
    \begin{table}[t]\centering\small
      \setlength{\tabcolsep}{7pt}\renewcommand{\arraystretch}{1.12}%
      {\centering\TprobeTab\par}
      \vspace{5pt}
      \TprobeCaption
    \end{table}%
  \fi}

\newcommand{\Fteaser}{%
\begin{figure}[t]
  \centering
  \includegraphics[width=0.7\linewidth]{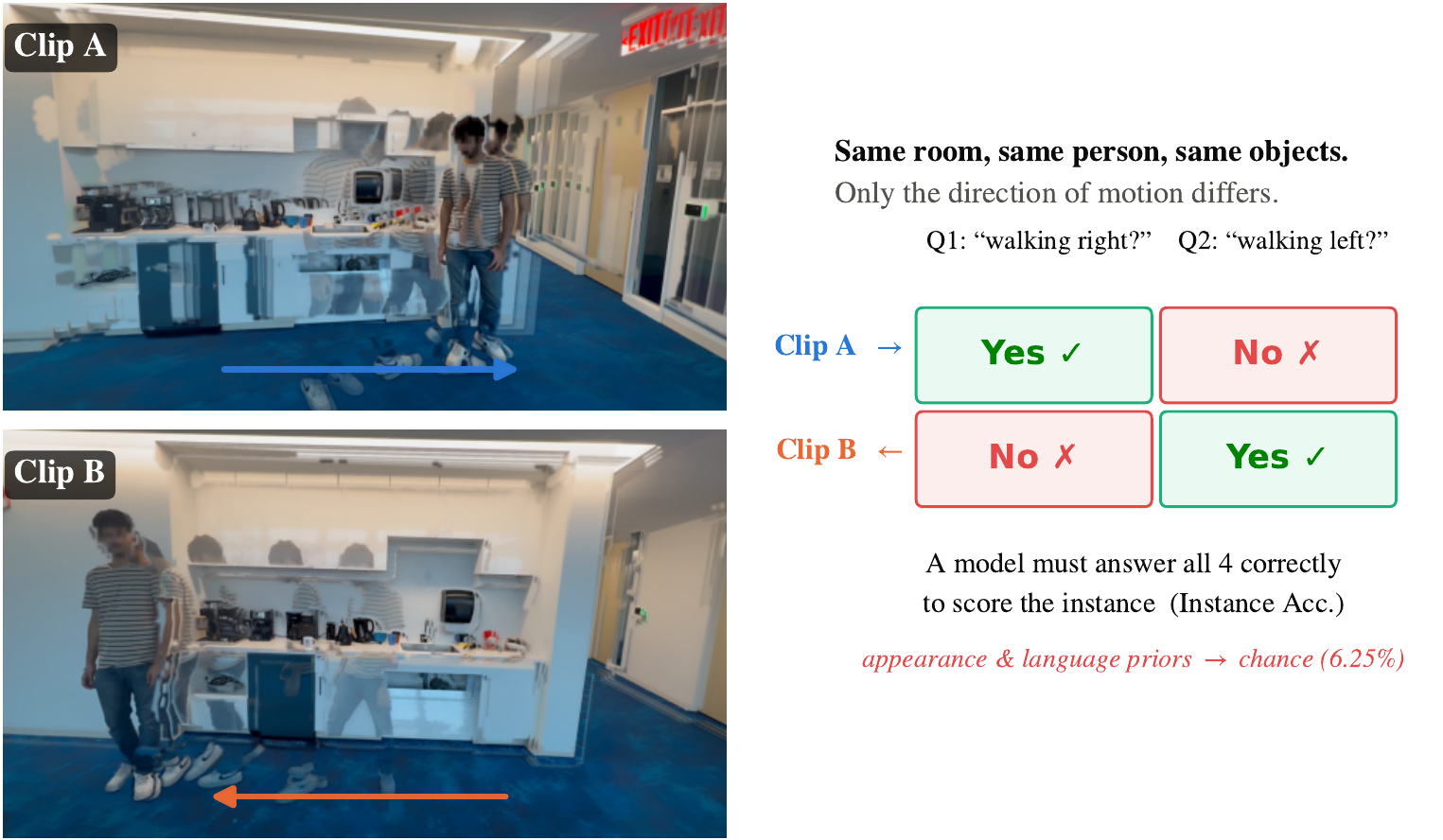}
  \caption{The illusion of motion understanding. A contrastive MotionBlind
    instance: two self-recorded clips share the same room, person, and objects and
    differ \emph{only} in the direction of motion (frames overlaid stop-motion
    style, faint$\rightarrow$solid, to visualize the trajectory). Each clip carries two complementary yes/no questions, and a model scores the instance only if all four items are correct (\iacc{}). Any strategy that leans
    on static appearance or language priors (naming the objects, guessing the
    likely action) is driven to the $6.25\%$ chance floor.}
  \label{fig:teaser}
\end{figure}}

\newcommand{\Fdataset}{%
\begin{figure}[t]
  \centering
  \includegraphics[width=0.90\linewidth]{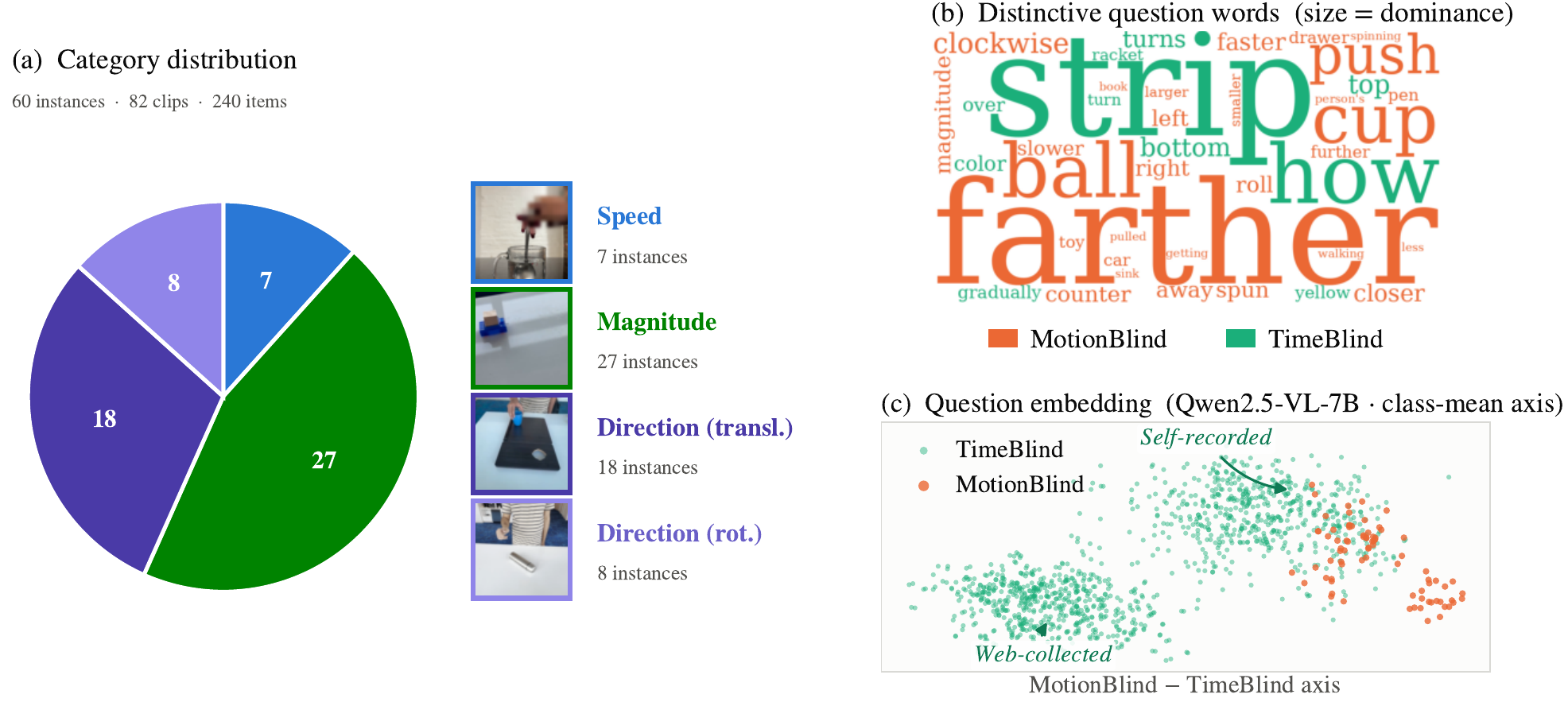}
  \caption{MotionBlind versus TimeBlind.
    \textbf{(a)} The self-recorded set: $60$ contrastive instances ($82$ clips, $240$
    items) across three physical-motion characteristics---speed, magnitude, and
    direction---each with a representative, face-blurred frame.
    \textbf{(b)} Words most distinctive of each benchmark's questions (weighted log-odds
    \citep{monroe2008fightin}): MotionBlind turns on manner, magnitude, and direction (\emph{faster, farther, closer, clockwise}); TimeBlind on appearance and state
    change (\emph{color, turns, top/bottom, gradually}). Manner terms appear in $94\%$ of
    MotionBlind questions versus $36\%$ of TimeBlind's.
    \textbf{(c)} The same questions in the evaluated models' backbone (Qwen2.5-VL-7B,
    mean-pooled), projected onto the axis between the two benchmarks' means. The two are
    linearly separable (AUC $0.97$); TimeBlind further splits into a self-recorded and a
    web-collected cluster, with MotionBlind adjacent to former.}
  \label{fig:dataset}
\end{figure}}
\newcommand{\Fsamplingdynamic}{%
\begin{figure*}[t]
  \centering
  \includegraphics[width=\textwidth]{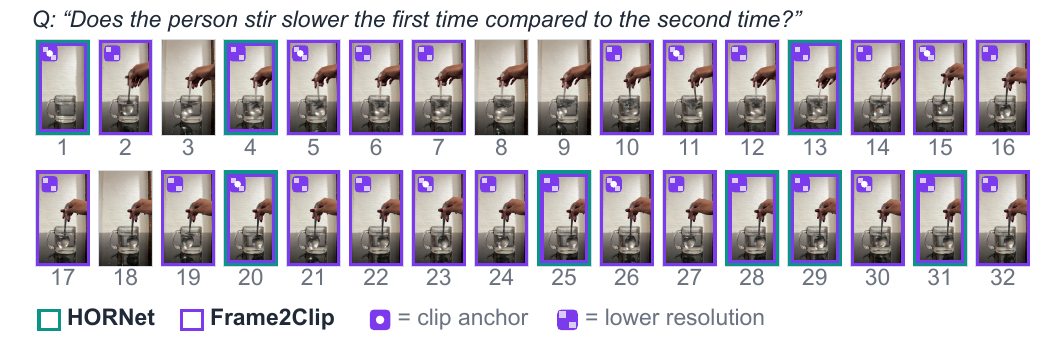}
  \caption{%
    Dynamic frame selection on one MotionBlind clip at a budget of $K=8$.
    The 32 candidate frames are spaced evenly over the clip; the question is
    ``Does the person stir slower the first time compared to the second time?''
    (ground truth: no).
    A colored border marks a frame that a selector returns.
    HORNet (teal) keeps the 8 frames with the highest keep probability.
    Frame2Clip (violet) places anchors at the 8 peaks of the SigLIP2 frame--query
    similarity (white dot) and grows each anchor into a contiguous clip rendered at
    decreased resolution (checkerboard), so 29 frames cost 7.25 of the 8 native-frame
    units.
  }
  \label{fig:sampling_dynamic}
\end{figure*}
}

\newcommand{\FbudgetCaption}{%
  \caption{More frames help, then plateau far below solving. MotionBlind \iacc\ vs.\ frame budget $N$ under each frame-selection strategy: (a) uniform, (b) random, (c) HORNet, (d) Frame2Clip. The dashed line marks chance ($6.25\%$). Every open model saturates by $N{=}16$--$24$ and plateaus at or below $12\%$ for every selector. In (d), F2C is flat by construction: F2C's $1$\,FPS pool saturates on second-scale clips (Sec.~\ref{sec:results}).}%
  \label{fig:budget}}

\newcommand{\Fsweeps}{%
  \begin{figure}[t]\centering
    \includegraphics[width=0.95\linewidth]{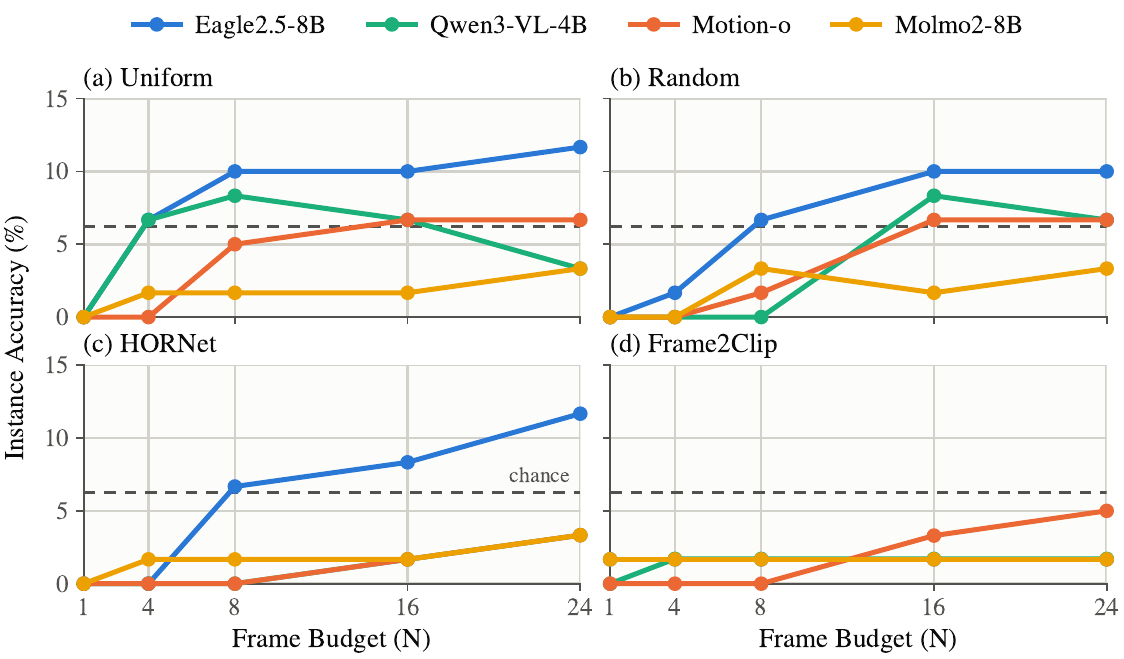}
    \FbudgetCaption
  \end{figure}}

\newif\ifcombinefigures
\combinefigurestrue

\newcommand{\Fbudget}{\Fsweeps}
\newcommand{\Fordercollapse}{}

\title{MotionBlind: Probing the Illusion of Motion Understanding in Video-LLMs}

\author{%
  Dhairya Bhatia\thanks{Equal contribution.} \And
  Bishoy Galoaa\footnotemark[1] \And
  Oliver Fritsche \And
  Shahid Kamal \AND
  Muhammad Obaidullah Abdul Salam \And
  Umer Saleem \And
  Om Rastogi \And
  Frania Felix Chettiar \And
  Nesli Erdo\u{g}mu\c{s} \And
  Sarah Ostadabbas\thanks{Corresponding author: \texttt{s.ostadabbas@northeastern.edu}} \\
  \AND
  Electrical and Computer Engineering Department \\ Northeastern University, Boston, MA, USA
}

\begin{document}

\maketitle

\begin{abstract}
Video large language models (Video-LLMs) are increasingly used as the perceptual front
end of world models, a role that assumes they can read \emph{motion}: how fast something
moves, which way it travels, how hard it is pushed. We show they cannot. A Video-LLM can
watch two clips of the same person in the same room, name every object in both, and still
fail to say which clip moves faster.
We introduce \textbf{MotionBlind}, a contrastive benchmark of self-recorded video for
physically grounded motion (speed, magnitude, and direction), the variables a world
model must predict. Each instance is a pair of near-identical clips that differ only in
motion. Each clip carries two complementary yes/no questions, giving four items per
instance, and a model earns credit only if all four are correct. We report Instance
Accuracy (\iacc), which has a $6.25\%$ chance floor. Single-frame, appearance, and
language-only shortcuts all collapse to it. MotionBlind complements the recent TimeBlind
benchmark.
We run a controlled study of six open and two frontier Video-LLMs, varying whether the
video is present, whether frames are shown in the correct temporal order, and how frames
are sampled ($1$ to $24$ frames, four selection strategies).
Open models sit near the $6.25\%$ floor, and scale does not help. Removing the video
drops every model to zero \iacc, and shuffling frames collapses \iacc\ to chance, so the
task genuinely needs video in order. Neither more frames nor smarter frame selection
closes the gap, because these change \emph{which} frames are seen, not whether motion is
read. Only Gemini~3.1~Pro clears the benchmark overall, and even it fails on speed. A
front end that cannot tell two speeds of the same action apart is not yet a trustworthy
source of supervision, reward, or evaluation for a world model.
\end{abstract}

\begin{center}
\small 
\href{https://huggingface.co/datasets/augmentedcognitionlab/MotionBlind}%
{\raisebox{-0.15em}{\includegraphics[height=1.2em]{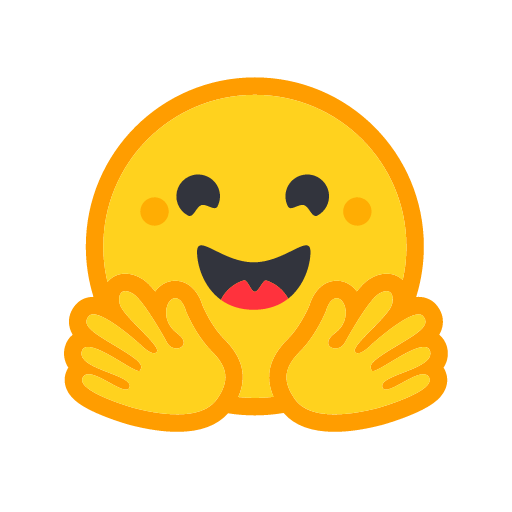}}~\textbf{MotionBlind Dataset}}%
\quad
\href{https://github.com/ostadabbas/MotionBlind}{\faGithub\ \texttt{ostadabbas/MotionBlind}}%
\quad
\href{https://ostadabbas.github.io/motionblind.github.io/\#}{\faGlobe\ \textbf{Project Page}}
\end{center}

\section{Introduction}
\label{sec:intro}

\Fteaser

Show a video large language model (Video-LLM) two short clips of the same person in
the same room, differing only in that one walks left and the other walks right, and it
will describe both scenes fluently (the person, the clothing, the furniture) and
then, as often as not, get the direction wrong. The static content is read perfectly;
the motion is not. This is the pattern that our MotionBlind benchmark is built to expose (Figure~\ref{fig:teaser}), while striking at a role Video-LLMs are increasingly asked to play.

World models~\citep{ha2018world} predict how an environment evolves, and increasingly they lean on large multimodal models rather than physics engines alone. Video-LLMs now sit across this stack: auto-labeling state transitions for data curation, acting as reward or success detectors in model-based reinforcement learning, and grounding language instructions in an agent's observations. Each role asks the model not to recognize \emph{what} is in a scene but to perceive \emph{how} it changes, direction, speed, duration, and the fine-grained kinematics a world model must ultimately predict.

This is a strong assumption, and a growing body of evidence suggests it does not hold. Benchmarks built around single video clips and free-form or multiple-choice questions consistently overestimate temporal competence, because a model can often answer correctly from a single representative frame, from object co-occurrence statistics, or from language priors about what typically happens next, without ever attending to motion \citep{liu2024tempcompass, moezziposition}. The fix that has emerged in recent work is a contrastive or minimal-pairs protocol, inspired by Winoground's visio-linguistic compositionality test for images \citep{thrush2022winoground}: construct two clips that share identical static content and differ only in temporal structure, pair each with two complementary questions, and require a model to get all four combinations right. Vinoground \citep{zhang2024vinoground} applied this idea to natural short videos and found that even GPT-4o barely exceeds chance on the strict group-level metric. Most directly relevant to us, the TimeBlind benchmark \citep{li2026timeblind} scales this paradigm to 600 curated instances (2,400 video-question pairs) spanning three levels of temporal composition (atomic events, event properties, and event interdependencies) and shows that the best of over twenty frontier  Multimodal Large Language Models (MLLMs)  reaches only 48.2\% Instance Accuracy, \iacc\footnote{\iacc{} is an instance-level accuracy metric: an instance is counted as correct only if all four predicted items are correct. This makes it a strict end-to-end evaluation metric with a chance level of $6.25\%$. Unless otherwise noted, \iacc{} is our primary evaluation metric. For reference, humans achieve 98.2\% on this metric on TimeBlind benchmark \citep{li2026timeblind}.}. Related diagnostic work shows the same failure mode along other motion axes, including signed motion direction \citep{lee2026directional} and camera motion \citep{feng2026geometry}, reinforcing that this is a perceptual bottleneck rather than a benchmark artifact.


TimeBlind's taxonomy, however, emphasizes semantic/compositional reasoning (state transitions, causal contingency, ordering). Two questions remain for physical AI: does the same collapse occur on physically-grounded properties (e.g., speed, magnitude, direction) isolated as controlled variables, and can anything be recovered purely by controlling which and how many frames a model sees, a lever every ingestion pipeline already has?

We introduce \textbf{MotionBlind}, a benchmark of $240$ video-question pairs ($60$ contrastive $2{\times}2$ instances over $82$ self-recorded clips) targeting physically grounded motion; speed, magnitude, and direction; that is hard to source and label from internet video but directly controllable in a lab (Figure~\ref{fig:dataset}). Where TimeBlind maps the breadth of temporal blindness, MotionBlind trades scope for controlled depth on the physical core of motion.

\Fdataset

Two design choices make the benchmark a clean diagnostic of motion understanding. First, all clips are self-recorded, with the objects, scene, and background held constant within each pair, ensuring that the only meaningful difference is the motion property under evaluation. Second, every relative-motion question compares two moving references within the same scene rather than relying on absolute concepts such as "fast," "slow," "hard," or "soft." This eliminates the need for context-dependent (operational) definitions, whose meanings vary with the scenario. For example, a speed considered slow for a moving car may be fast for a walking person.

We benchmark Video-LLMs on MotionBlind alongside two reference datasets: \textbf{TimeBlind} \citep{li2026timeblind}, which shares our contrastive $2{\times}2$ protocol but only targets semantic temporal reasoning rather than isolating physical motion (speed, direction, magnitude) as a controlled variable, and \textbf{Video-MME} \citep{fu2024videomme}, a standard non-contrastive video-QA suite that establishes the models we test are broadly capable and calibrates what our integrity probes do to a benchmark that does \emph{not} require motion understanding. Our controlled diagnostic study evaluates Video-LLMs along three complementary axes. First, we assess video necessity using a no-video, text-only probe to determine whether the questions can be answered without visual input. Second, we evaluate temporal-order sensitivity using shuffled- and reversed-frame probes, which preserve the visual content of individual frames while disrupting their temporal order. Third, we study the effect of frame selection by comparing four strategies across frame budgets ranging from 1 to 24 frames: uniform sampling, random sampling, a GRPO-trained learned frame selector (HORNet \citep{bai2026hornet}, trained with GRPO \citep{shao2024deepseekmath}), and a training-free key-clip selector (Frame2Clip \citep{sun2025frames2clips}). Together, these experiments test whether motion-centric questions can be solved without video, without correct temporal ordering, from a single frame, by increasing the number of sampled frames, or by employing more sophisticated frame-selection strategies.

We evaluate six open Video-LLMs spanning different design philosophies: Eagle2.5-8B \citep{chen2025eagle}, a long-context model; Qwen3-VL-4B \citep{qwen2025vl3}, a recent general-purpose model; Motion-o (7B) \citep{galoaa2026motiono}, a motion-tuned variant of Qwen2.5-VL-7B \citep{qwen2025vl}; Gemma-4-12B-it \citep{gemmateam2026gemma4}, a larger general-purpose model; Molmo2-8B \citep{ai2025molmo2,deitke2024molmo}, an open video-native model; and Inkling-Small \citep{inkling2026}, a reasoning model. We further compare against two frontier proprietary models, Gemini~3.1~Pro \citep{gemini31pro2026} and GPT-5.6 Luna \citep{gpt56luna2026}.

On MotionBlind, the task genuinely requires video in the correct order: text-only accuracy is exactly zero, and shuffling or reversing frames collapses \iacc\ toward chance. Every open model stays near that floor regardless of scale, and neither more frames nor dynamic frame selection (learned or training-free) beats plain uniform sampling at matched budgets; adaptive selection changes \emph{which} frames are seen, not whether model inherently understands the motion presented in a video sequence. Only the frontier Gemini~3.1~Pro clears MotionBlind overall, and even it fails on the purely rate-defined categories, isolating a representational gap that sampling strategies do not close.

We position MotionBlind as both a diagnostic benchmark and, ultimately, a training signal for the perceptual front end of Physical AI world models. A system that cannot reliably distinguish two visually identical actions performed at different speeds, magnitudes, or directions cannot serve as a trustworthy source of physically grounded supervision, reward, or evaluation. Our contributions are:

\begin{itemize}
    \item MotionBlind, a contrastive, self-recorded benchmark for evaluating physically grounded motion perception through three fundamental motion attributes: speed, magnitude, and direction. MotionBlind complements TimeBlind's semantically driven taxonomy while adopting the same strict \iacc{} evaluation metric.
    
    \item A controlled diagnostic study that systematically isolates three factors in video understanding: (i) the necessity of video, (ii) sensitivity to temporal order, and (iii) frame-selection strategy. Across six open Video-LLMs, two frontier proprietary models, and frame budgets from $1$ to $24$, we show that the observed failures are fundamentally perceptual and order-dependent rather than purely linguistic, and persist even as model capability scales.
    
    \item An analysis of frame selection showing that neither a GRPO-trained learned selector nor a training-free key-clip selector consistently outperforms uniform sampling for pairwise motion discrimination. Our analysis reveals that current frame selection methods largely ignore temporal ordering and produce nearly uniform frame-importance scores, motivating future order-aware and discrimination-aware training objectives.
\end{itemize}

\section{Related Work}
\label{sec:related}

\paragraph{Video-LLMs and their perceptual front ends.} Modern Video-LLMs sample a fixed budget of frames, encode them with a vision tower, and condition a language model for captioning and QA. This "frame aggregation + LLM reasoning" paradigm drives gains on aggregate benchmarks \citep{moezziposition}, exemplified by long-context sampling in Eagle2.5-8B, the Qwen-VL line's dynamic resolution \citep{qwen2025vl,qwen2025vl3}, and open video-native models such as Molmo/Molmo2 \citep{deitke2024molmo,ai2025molmo2}. We treat these as fixed front ends and probe what they perceive once static appearance is removed, rather than proposing an architecture.  To isolate that residual, recent benchmarks use minimal pairs of near-identical videos (TempCompass \citep{liu2024tempcompass}, Vinoground \citep{zhang2024vinoground,thrush2022winoground}, TimeBlind \citep{li2026timeblind}); we adopt TimeBlind's \iacc\ metric and $2{\times}2$ protocol and extend it toward physically grounded, controllable properties (speed, magnitude, direction).


\paragraph{Frame sampling and selection for Video-LLMs.} Which frames a model sees strongly affects aggregate scores such as Video-MME \citep{fu2024videomme}. Beyond uniform and random sampling, learned selectors score frames by informativeness for single-answer QA (e.g.\ HORNet~\citep{bai2026hornet}, trained with GRPO), while training-free methods select temporally coherent \emph{clips} to preserve motion (e.g.\ Frame2Clip \citep{sun2025frames2clips}). We evaluate both and find their long-form gains do not transfer to pairwise, order-sensitive discrimination; a consequence of selectors whose reward and encoders carry no notion of temporal position or of the paired video.

World models predict how an environment evolves for planning and imagined rollouts \citep{ha2018world,hafner2020dreamer}, and increasingly use pretrained multimodal models as data generators, labelers, or plausibility evaluators. The reliability of that component's motion perception is therefore a precondition for the world model's physical grounding, and MotionBlind is a targeted diagnostic for exactly this precondition~\citep{agarwal2025cosmos, nvidia2026cosmos3}.


\section{Introducing MotionBlind}
\label{sec:dataset}

MotionBlind is a self-recorded contrastive benchmark designed to isolate physically-grounded motion understanding from appearance, object identity, scene context, and language priors. Clips are recorded at $30$\,fps and run $5.9$\,s on average (median $4.5$\,s,
range $1.9$--$15.7$\,s), at resolutions from $480{\times}848$ to
$1280{\times}1280$, predominantly portrait. Unlike existing video benchmarks, every example is constructed so that the only meaningful difference between two clips is the motion being performed. Consequently, the correct answer cannot be inferred from a single frame, object recognition, or semantic context.

MotionBlind contains 60 contrastive instances (82 video clips and 240 question-answer items, with answers balanced 50/50 between yes and no). Each instance consists of a minimal pair of near-identical videos that share the same object, scene, background, camera viewpoint, and lighting while differing only in one motion attribute. By controlling all other visual cues, MotionBlind ensures that motion is the only signal capable of determining the correct answer.

The benchmark focuses on three fundamental physical properties of motion: \emph{speed} (7 instances), \emph{magnitude} (27 instances), and \emph{direction} (26 instances, including 18 translational and 8 rotational motions) (see Figure~\ref{fig:dataset}a). The paired clip serves only to define the comparison structure and is never required to determine the answer to a question.

\paragraph{Comparison Benchmarks.}
We evaluate MotionBlind alongside TimeBlind \citep{li2026timeblind}, which adopts the same contrastive $2\times2$ evaluation protocol but targets semantic temporal reasoning rather than physically grounded motion. TimeBlind contains 600 contrastive instances (2,400 question-answer items) collected from curated internet videos, simulations, and real-world recordings. While both datasets share an identical evaluation protocol, they probe complementary aspects of video understanding. MotionBlind asks \emph{how} an action is performed by focusing on speed, magnitude, and direction, whereas TimeBlind asks \emph{what} changes and \emph{when}. This distinction is reflected in the language of the datasets, with motion-related terms appearing in 94\% of MotionBlind questions compared with 36\% of TimeBlind questions, and in the embedding space of Qwen2.5-VL-7B, where the two datasets are highly separable (AUC 0.97; Figure~\ref{fig:dataset}b,c).

As a non-contrastive reference, we also evaluate on Video-MME \citep{fu2024videomme}, which contains 2,700 four-way multiple-choice questions over 900 internet videos and is scored using standard item accuracy. Video-MME measures general video question answering ability, whereas MotionBlind and TimeBlind specifically isolate temporal reasoning through contrastive evaluation.

\paragraph{Evaluation Metric.}
MotionBlind adopts the same evaluation protocol as TimeBlind \citep{li2026timeblind}, which extends the contrastive evaluation framework introduced by Winoground \citep{thrush2022winoground}. Each instance forms a $2\times2$ contrastive task consisting of two near-identical videos ($i_0$, $i_1$) and two complementary questions ($q_0$, $q_1$). The ground-truth answers follow a diagonal pattern: $q_0$ is true for $i_0$ and false for $i_1$, while $q_1$ is false for $i_0$ and true for $i_1$, yielding four binary prediction items per instance.

Because the paired videos share identical appearance, objects, and scene context while differing only in the motion attribute under evaluation, a model that ignores temporal information cannot satisfy the complete $2\times2$ pattern, regardless of how strong its object recognition or language priors are.

We report four nested evaluation metrics of increasing strictness. \acc{} measures the fraction of correctly answered items independently (chance $50\%$) and is the most permissive metric. \qacc{} and \vacc{} require both items associated with a question or a video, respectively, to be answered correctly (chance $25\%$). Our primary metric is \iacc{}, which counts an instance as correct only if all four prediction items are correct simultaneously (chance $6.25\%$). This strict instance-level metric measures complete understanding of the contrastive motion relationship and prevents models from receiving partial credit by exploiting appearance or language biases.
\paragraph{Human Evaluation.}
To establish the human ceiling and verify independent answerability, we conducted an internal study with five annotators. Each annotator evaluated all 60 instances. Crucially, clips were presented \emph{one at a time} (not side-by-side) to ensure that the ground-truth answer for each clip could be determined from its motion alone, without requiring direct visual comparison to its paired clip. Annotators achieved a mean \iacc{} of 91.3\% (range 85--97\%) and a per-item \acc{} of 97.8\%, confirming that the benchmark is trivially solvable by humans when the video is present and ordered.

\section{Experimental Setup}
\label{sec:setup}

\paragraph{Models.} We evaluate six open Video-LLMs chosen to span training emphases:
Eagle2.5-8B, a long-context model optimized for processing high-frame-count video sequences; Qwen3-VL-4B, a recent general model that encodes time as
textual timestamp tokens prefixed to each temporal patch; Motion-o (7B), a motion-tuned
variant of Qwen2.5-VL-7B, whose backbone instead ties temporal position
IDs to absolute time; Gemma-4-12B-it, a larger general
model with a native video path; Molmo2-8B, an open video-native model; and
Inkling-Small, a reasoning model (evaluated with reasoning disabled to match
the benchmark's no-CoT convention). We additionally evaluate two frontier proprietary models (Gemini 3.1 Pro, GPT-5.6 Luna) as reference points. All models are treated as fixed perceptual front ends; we do not
fine-tune.

\paragraph{Controlled axes.}
We vary three levers while holding the model fixed.
\emph{(i) Frame budget}: $N\in\{1,4,8,16,24\}$ frames. \emph{(ii) Frame selection}:
uniform (evenly spaced), random, HORNet (a GRPO-trained learned
selector), and Frame2Clip (training-free key-clip selection); for HORNet we use the
released short-video checkpoint. \emph{(iii) Integrity probes} at a matched $16$-frame
budget: \emph{no-video} text-only, \emph{shuffled}-frames (content
preserved, order destroyed), and \emph{reversed}-frames.

\paragraph{Implementation Details.} We append \textit{``Please output Yes or No.''} to every yes/no question (for
TimeBlind's multiple-choice items, \textit{``Please output A or B.''}) and parse an explicit \textit{``Answer: yes/no''} (resp.\ \textit{A/B}) or, failing that, the first standalone \textit{yes}/\textit{no} (\textit{A}/\textit{B}); unparseable
outputs are scored incorrect. For Video-MME we instead use its standard four-way prompt (\textit{``Respond with only the letter\dots''}) and extract the chosen option with ordered patterns, under the same unparseable-scores-wrong rule. All models use greedy decoding ($T{=}0.0$). Molmo2-8B and Qwen3-VL-4B run the default zero-shot prompt at $128$ tokens; Eagle2.5-8B and Motion-o, which degrade under constrained generation, use their authors' recommended chain-of-thought scaffolds ($300$ tokens, repetition penalty $1.2$).

\section{Experimental Results}
\label{sec:results}

Across every model, frame budget, and sampling strategy we test, motion understanding is not solved, and the levers that help most are the ones that confirm, rather than close, the gap.

\paragraph{Open models fail; only a frontier model breaks away.}
Table~\ref{tab:main} reports the headline numbers. Among open models the picture is
uniform: the best, Eagle2.5-8B, reaches only 11.7\% \iacc{} on MotionBlind (Qwen3-VL-4B, strongest on TimeBlind at 24.7\%, manages 8.3\%); the rest span 10.0\% down to 3.3\%, with a larger 12B parameter model (Gemma-4-12B-it) landing among the weakest at 3.6\% on the 56-instance subset, so scale alone does not help. Only the frontier Gemini~3.1~Pro breaks away, at
$60.0\%$ \iacc\ ($80.8\%$ per-item \acc); GPT-5.6 Luna stays at $15.0\%$, near the
open-model band. Though Gemini performs better relative to other tested models, it still remains well short of the human ceiling ($91.3\%$ \iacc,
$97.8\%$ \acc; five annotators) and, as we show below, collapses on the hardest physical
categories. For the open models \acc\ sits in the $52$--$59\%$ band even when \iacc\ is at the floor, certifying fluent guessing on a
balanced yes/no set rather than the ability to tell the two clips apart. This gap
between \acc\ and \iacc\ is the illusion the benchmark is named for (the open models we test on it reach $58$--$70\%$ on Video-MME in their best configuration, Table~\ref{tab:main}; full budget in Figure~\ref{fig:sweep-vmme} and Table~\ref{app:sweep-vmme}, Appendix~\ref{app:videomme}).

\Tmain

\paragraph{The task genuinely requires video and temporal order.}
A key question is whether models succeed by understanding motion or by exploiting appearance and language shortcuts. To answer this, we evaluate three integrity probes across four models and three datasets: (i) shuffled frames, (ii) reversed frames, and (iii) a text-only setting in which the video is removed entirely. The first two probes preserve the visual content of individual frames while disrupting temporal relationships, whereas the third tests whether the prompted questions can be answered from language alone. Table~\ref{tab:probe} shows that MotionBlind is highly dependent on both video content and correct temporal ordering. Removing the video or disrupting frame order reduces every model's \iacc{} to at or near $0\%$, indicating that the benchmark cannot be solved reliably without reasoning over motion across frames.
The same probes produce a very different outcome on the comparison benchmarks. On Video-MME, shuffling frames reduces \acc{} by only $0.4$ to $1.9$ points, reversing frames by $1.0$ to $2.5$ points, and removing the video entirely still leaves models with $39$ to $44\%$ accuracy, substantially above the $25\%$ chance level. This suggests that many Video-MME questions can be answered from appearance and language priors alone. TimeBlind behaves differently: removing the video reduces performance to near $0\%$ \iacc{}, confirming that the benchmark cannot be solved from language alone, while shuffled and reversed frames still yield approximately chance-level performance ($6.25\%$), indicating that some instances remain solvable without preserving the correct temporal order.

Together, these results show that MotionBlind simultaneously eliminates language leakage and enforces temporal reasoning. Success requires both the video and an understanding of the temporal relationships between frames, making motion perception a prerequisite for solving the benchmark.
\Tprobe
\Fordercollapse

\paragraph{The gap is concentrated in physical magnitude.}
Breaking \iacc\ down by category (Figure~\ref{fig:category}) localizes the failure. For
every open model, \emph{speed} and \emph{magnitude} sit at zero \iacc{}; only
\emph{direction (translational)} carries any signal ($22$--$33\%$). Gemini~3.1~Pro is the
exception, clearing chance on three of four categories, yet it too drops to $14\%$ on
\emph{speed}, the one quantity defined purely by rate. The models are not uniformly blind
to time; they specifically fail to encode how fast and how far, exactly the kinematics a
world model must predict.

\paragraph{More frames help, then plateau far below solving.}
\Fbudget
\vspace{-5pt}
Since the models are order-sensitive, does simply showing more of the clip help?
Figure~\ref{fig:budget} sweeps the frame budget under each selection strategy. Accuracy rises
with $N$ and saturates around $N{=}16$--$24$ for every model, but the ceiling is low:
Eagle plateaus near 12\% and the others at or below $\sim$8\%. Frames are a lever with a hard
stop: they let the model see more of a motion it still cannot represent.

\paragraph{Dynamic frame selection does not rescue it.}
If uniform sampling saturates, perhaps a smarter selector would spend the budget
better. It does not. On the flagship Eagle model, the
GRPO-trained learned selector (HORNet) \emph{underperforms} plain uniform sampling at a
matched $16$--frame budget ($10.0$ uniform vs.\ $8.3$ HORNet on MotionBlind), while
random sampling merely matches it; HORNet occasionally nudges a weaker
model's number up or down, but no strategy pushes Eagle above $11.7\%$ at any budget (with Molmo2 never above $3.3\%$). The learned selector
offers no lever here for a diagnosable reason: on this set its frame keep-probabilities
are nearly flat (near-zero variance on $76$ of $82$ clips), so its top-$K$ picks are
close to arbitrary rather than a tuned, motion-aware selection; Figure~\ref{fig:sampling_dynamic} shows which frames each selector keeps and full-budget in Figure~\ref{fig:sweep-mb}. Following \citet{sun2025frames2clips}, we pool F2C candidates at $1$\,FPS.
MotionBlind clips average $5.9$\,s, so the pool is smaller than
$K_{\mathrm{anchor}}$ on $73\%$ of clips at $N{=}8$ and on all $82$ at
$N{=}24$: F2C returns the full pool rather than a selection, and the anchor
sets at $N{=}16$ and $N{=}24$ are identical. The F2C column is therefore a
scope boundary rather than evidence about frame selection, and we report it
as such. Changing which frames are shown does not change whether motion is read (all samplers $\times$ budgets, with exact numbers, in Table~\ref{app:sweep-mb}.

\section{Discussion: Implications for World Models}
\label{sec:discussion}

The through-line is that no post-hoc lever (more frames, smarter selection, better
prompting) recovers a signal the representation never encoded, and even the one frontier
model that clears MotionBlind overall still fails on physical magnitude. This matters for
the world model stack: a Video-LLM used to auto-label transitions, score success for
model-based RL, or ground instructions is trusted to read exactly the motion it here
fails to read. A reward model that cannot tell a gentle placement from a forceful slam
supplies noisy or inverted supervision on the physical variables that matter most, and
because \acc\ looks healthy the failure is easy to miss. The $\sim$$19$-point text-only floor on Video-MME is the quantified form of this illusion: a benchmark can look healthy while the model reads no motion at all. Two implications follow:
pipelines that use learned keyframe selection to cut cost inherit the selector's blind
spots, removing the temporal coverage a physical predictor needs; and MotionBlind's
contrastive, order-sensitive structure is itself a training signal; a pairwise,
order-aware objective that rewards telling $i_0$ from $i_1$ targets the gap that
appearance-driven pretraining leaves open.

\textbf{6.1 Limitations}

Our study has limitations. MotionBlind's 60 self-recorded instances trade scale for controlled isolation of physical variables, so results carry small-sample variance; all clips feature a single actor indoors, leaving in-the-wild and egocentric generalization open; and constrained yes/no parsing, while removing free-form ambiguity, may understate reasoning a model cannot verbalize in binary form. Our open-weight models span $\le$12B parameters, so we cannot fully separate architectural from scale limits, though a larger model not closing the gap is suggestive.

\textbf{6.2 Broader Impact and Ethics}

MotionBlind is a diagnostic for failure modes that matter in safety-critical physical AI and robotics, where misreading force or speed could cause unsafe actions. All video was collected with the participant's informed consent, faces are automatically blurred prior to release, and the dataset contains no sensitive or personally identifiable information.

\section{Conclusion}
\label{sec:conclusion}

We introduced MotionBlind, a self-recorded contrastive benchmark isolating physically grounded motion perception, and used it with TimeBlind to study six open Video-LLMs and two frontier models. The conclusion is consistent: for open models motion understanding is not solved, the residual signal is genuinely temporal, and neither more frames, larger models, nor dynamic selection closes the gap; even the frontier model that clears the benchmark overall fails on physical magnitude. A front end that cannot tell two speeds of the same action apart is not yet a trustworthy foundation for a world model. Its taxonomy also specifies further conditions for future collection: mass, acceleration, and a world-frame setting where the camera tracks the object; probing whether the perception inference gap widens as the inferred quantity moves further from the visible motion.

\bibliographystyle{plainnat}
\bibliography{references}

\newpage
\appendix

\section{Per-category breakdown}\label{app:category}
\begin{figure}[h]
  \centering
  \includegraphics[width=0.5\linewidth]{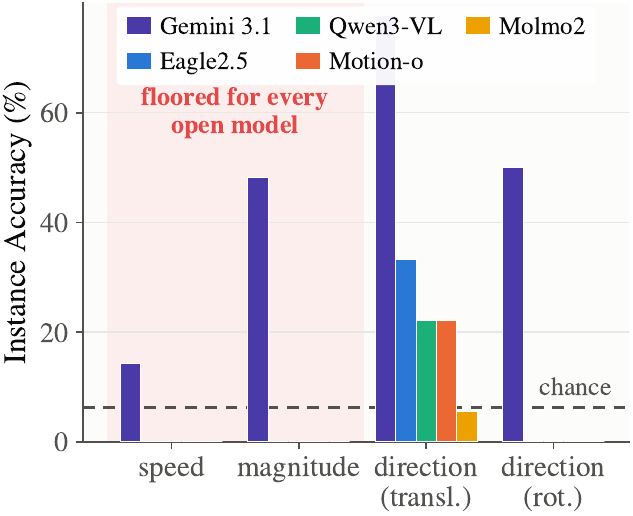}
  \caption{Per-category \iacc\ on MotionBlind (matched uniform-$16$ budget). For every open
    model, \emph{speed} and \emph{magnitude} are at the chance floor and only
    \emph{direction (translational)} carries signal; Gemini~3.1~Pro clears three of four
    categories yet still fails on \emph{speed}.}
  \label{fig:category}
\end{figure}

\section{Full frame-selection sweeps}\label{app:sweeps}
\begin{figure}[h]
  \centering
  \includegraphics[width=\linewidth]{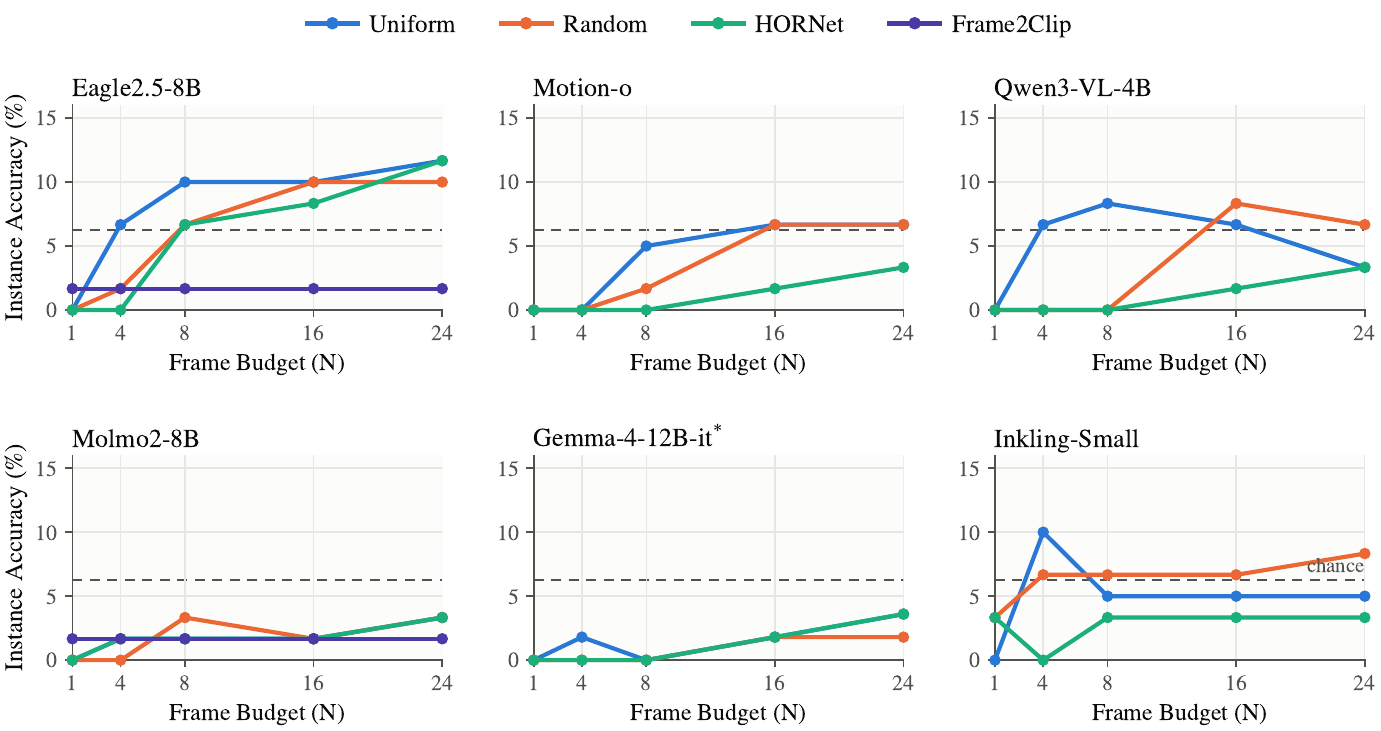}
  \caption{MotionBlind \iacc\ vs.\ frame budget $N$, one line per frame selector. Every open
    model plateaus in the low teens or below and no selector escapes the floor.
    $^{*}$Gemma-4-12B-it is on the MotionBlind-$82$ cache subset (not $1{:}1$ comparable).}
  \label{fig:sweep-mb}
\end{figure}

\begin{figure}[h]
  \centering
  \includegraphics[width=\linewidth]{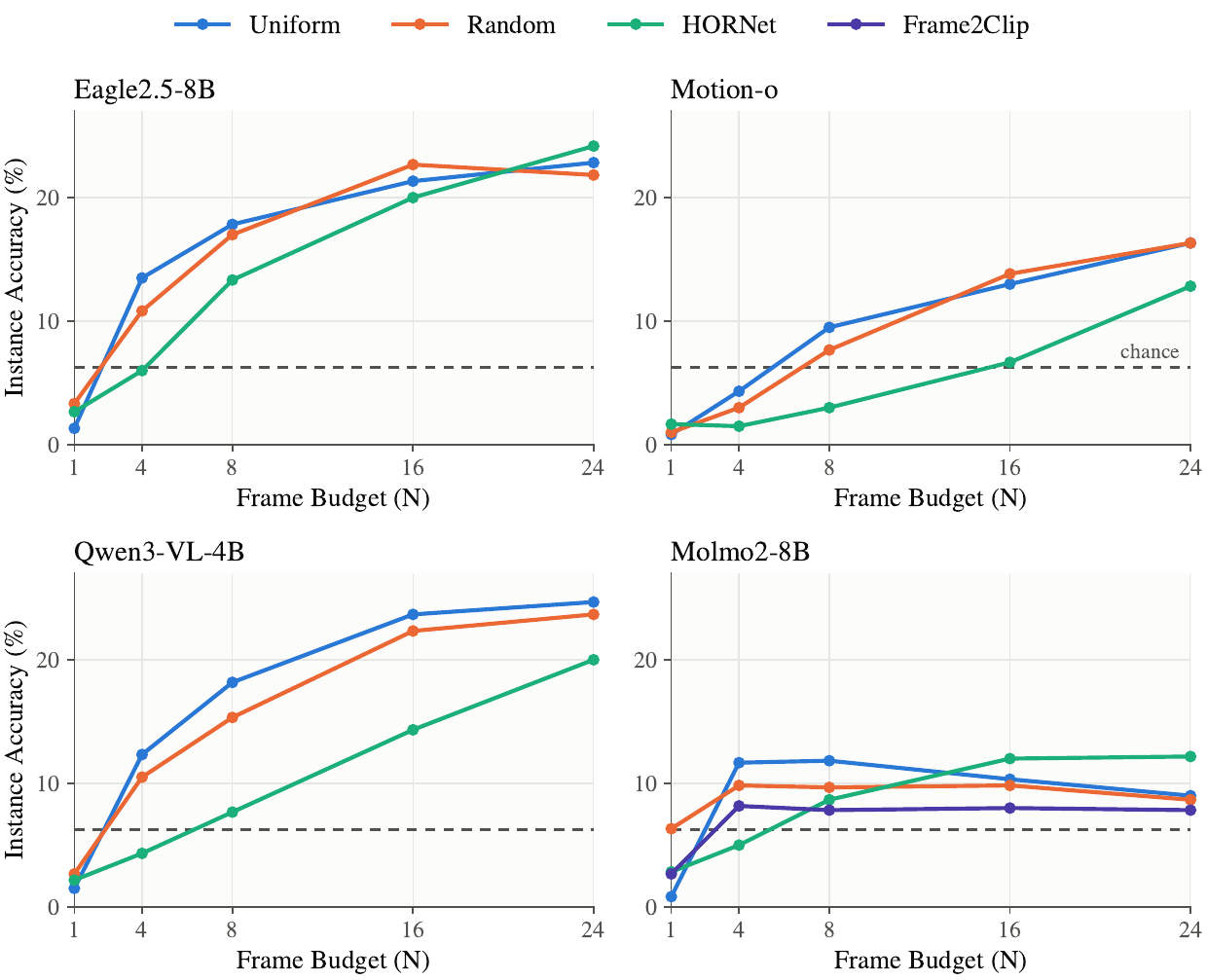}
  \caption{TimeBlind \iacc\ vs.\ frame budget $N$ (same models and selectors). Accuracy rises
    with $N$ but saturates well short of the human ceiling; Molmo2-8B adds a Frame2Clip curve.}
  \label{fig:sweep-tb}
\end{figure}

Per-selector budget sweeps for every open model on MotionBlind (Figure~\ref{fig:sweep-mb})
and TimeBlind (Figure~\ref{fig:sweep-tb}); all use the shared sampler key.

\Fsamplingdynamic{}

\section{Video-MME sanity check}\label{app:videomme}
Table~\ref{tab:probe} (Video-MME panel) shows why the non-contrastive reference matters: shuffling frames moves accuracy by only 0.4--1.9 points, reversing by 1.0--2.5, and removing the video entirely still leaves 39--44\% against the 25\% chance level. Standard multiple-choice video QA is therefore substantially answerable from appearance and language priors alone, whereas MotionBlind's contrastive protocol removes that leakage (text-only IAcc $=0$ for every model).
\begin{figure}[h]
  \centering
  \includegraphics[width=\linewidth]{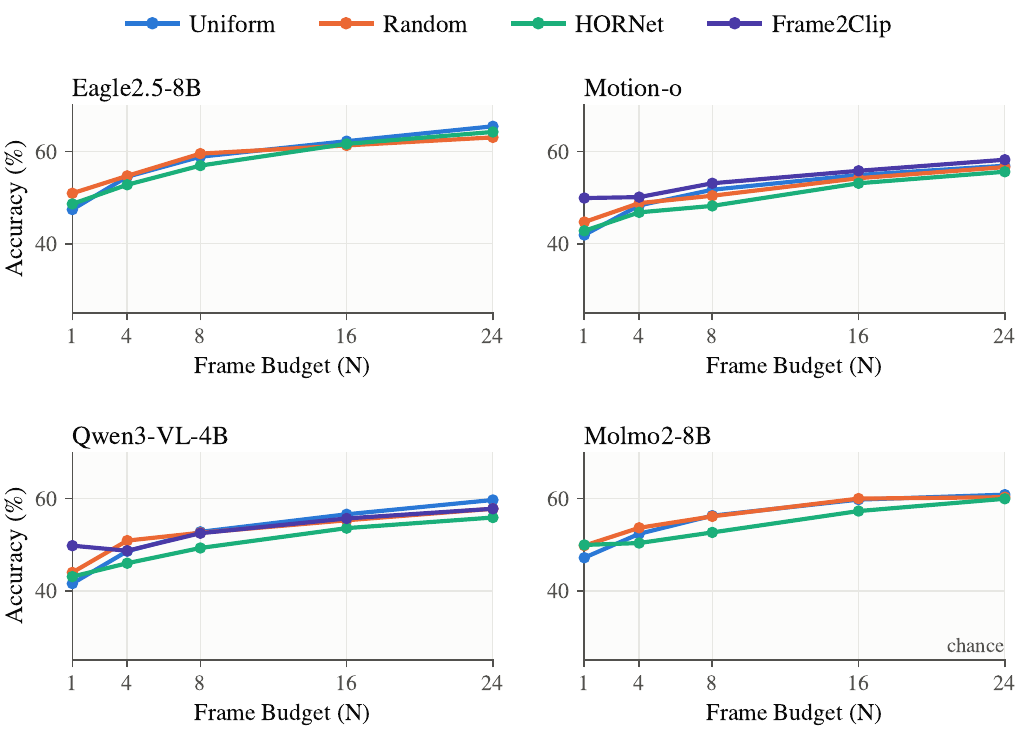}
  \caption{Video-MME accuracy vs.\ frame budget $N$ (chance $25\%$). The same models reach
    $\sim$$50$--$65\%$, so the MotionBlind failure is motion-specific rather than general
    incompetence. Eagle2.5-8B and Molmo2-8B are from our runs; Motion-o and Qwen3-VL-4B also
    include a Frame2Clip curve.}
  \label{fig:sweep-vmme}
\end{figure}

\section{Full sweep results (all numbers)}\label{app:sweep-tables}
The exact per-configuration values behind Figures~\ref{fig:sweep-mb}--\ref{fig:sweep-vmme}
and the frame-selection comparison: every model $\times$ frame selector $\times$ budget $N$, on
MotionBlind (Table~\ref{app:sweep-mb}), TimeBlind (Table~\ref{app:sweep-tb}), and Video-MME
(Table~\ref{app:sweep-vmme}).
\begin{table}[t]
  \centering\footnotesize
  \setlength{\tabcolsep}{6pt}\renewcommand{\arraystretch}{1.05}
  \caption{Full MotionBlind sweep: Instance Accuracy \iacc\ (\%) for every model, frame selector, and budget $N$ (chance $6.25\%$). ``--'' = not run. $^{*}$Gemma-4-12B-it on the MotionBlind-82 cache subset.}
  \label{app:sweep-mb}
  \begin{tabular}{@{}l ccccc@{}}
    \toprule
    Sampler & $N{=}1$ & $4$ & $8$ & $16$ & $24$ \\
    \midrule
    \multicolumn{6}{@{}l}{\textit{Eagle2.5-8B}}\\
    Uniform & 0.0 & 6.7 & 10.0 & 10.0 & 11.7 \\
    Random & 0.0 & 1.7 & 6.7 & 10.0 & 10.0 \\
    HORNet & 0.0 & 0.0 & 6.7 & 8.3 & 11.7 \\
    Frame2Clip & 1.7 & 1.7 & 1.7 & 1.7 & 1.7 \\
    \midrule
    \multicolumn{6}{@{}l}{\textit{Qwen3-VL-4B}}\\
    Uniform & 0.0 & 6.7 & 8.3 & 6.7 & 3.3 \\
    Random & 0.0 & 0.0 & 0.0 & 8.3 & 6.7 \\
    HORNet & 0.0 & 0.0 & 0.0 & 1.7 & 3.3 \\
    Frame2Clip & 0.0 & 1.7 & 1.7 & 1.7 & 1.7 \\
    \midrule
    \multicolumn{6}{@{}l}{\textit{Motion-o (7B)}}\\
    Uniform & 0.0 & 0.0 & 5.0 & 6.7 & 6.7 \\
    Random & 0.0 & 0.0 & 1.7 & 6.7 & 6.7 \\
    HORNet & 0.0 & 0.0 & 0.0 & 1.7 & 3.3 \\
    Frame2Clip & 0.0 & 0.0 & 0.0 & 3.3 & 5.0 \\
    \midrule
    \multicolumn{6}{@{}l}{\textit{Gemma-4-12B-it$^{*}$}}\\
    Uniform & 0.0 & 1.8 & 0.0 & 1.8 & 3.6 \\
    Random & 0.0 & 0.0 & 0.0 & 1.8 & 1.8 \\
    HORNet & 0.0 & 0.0 & 0.0 & 1.8 & 3.6 \\
    \midrule
    \multicolumn{6}{@{}l}{\textit{Molmo2-8B}}\\
    Uniform & 0.0 & 1.7 & 1.7 & 1.7 & 3.3 \\
    Random & 0.0 & 0.0 & 3.3 & 1.7 & 3.3 \\
    HORNet & 0.0 & 1.7 & 1.7 & 1.7 & 3.3 \\
    Frame2Clip & 1.7 & 1.7 & 1.7 & 1.7 & 1.7 \\
    \midrule
    \multicolumn{6}{@{}l}{\textit{Inkling-Small}}\\
    Uniform & 0.0 & 10.0 & 5.0 & 5.0 & 5.0 \\
    Random & 3.3 & 6.7 & 6.7 & 6.7 & 8.3 \\
    HORNet & 3.3 & 0.0 & 3.3 & 3.3 & 3.3 \\
    \bottomrule
  \end{tabular}
\end{table}

\begin{table}[t]
  \centering\footnotesize
  \setlength{\tabcolsep}{6pt}\renewcommand{\arraystretch}{1.05}
  \caption{Full TimeBlind sweep: Instance Accuracy \iacc\ (\%) (chance $6.25\%$).}
  \label{app:sweep-tb}
  \begin{tabular}{@{}l ccccc@{}}
    \toprule
    Sampler & $N{=}1$ & $4$ & $8$ & $16$ & $24$ \\
    \midrule
    \multicolumn{6}{@{}l}{\textit{Eagle2.5-8B}}\\
    Uniform & 1.3 & 13.5 & 17.8 & 21.3 & 22.8 \\
    Random & 3.3 & 10.8 & 17.0 & 22.7 & 21.8 \\
    HORNet & 2.7 & 6.0 & 13.3 & 20.0 & 24.2 \\
    \midrule
    \multicolumn{6}{@{}l}{\textit{Qwen3-VL-4B}}\\
    Uniform & 1.5 & 12.3 & 18.2 & 23.7 & 24.7 \\
    Random & 2.7 & 10.5 & 15.3 & 22.3 & 23.7 \\
    HORNet & 2.2 & 4.3 & 7.7 & 14.3 & 20.0 \\
    \midrule
    \multicolumn{6}{@{}l}{\textit{Motion-o (7B)}}\\
    Uniform & 0.8 & 4.3 & 9.5 & 13.0 & 16.3 \\
    Random & 1.0 & 3.0 & 7.7 & 13.8 & 16.3 \\
    HORNet & 1.7 & 1.5 & 3.0 & 6.7 & 12.8 \\
    \midrule
    \multicolumn{6}{@{}l}{\textit{Gemma-4-12B-it}}\\
    Uniform & 0.3 & 9.8 & 12.3 & 13.3 & 14.3 \\
    Random & 1.3 & 6.7 & 11.7 & 14.0 & 13.3 \\
    HORNet & 1.5 & 3.7 & 7.0 & 11.0 & 13.5 \\
    \midrule
    \multicolumn{6}{@{}l}{\textit{Molmo2-8B}}\\
    Uniform & 0.8 & 11.7 & 11.8 & 10.3 & 9.0 \\
    Random & 6.3 & 9.8 & 9.7 & 9.8 & 8.7 \\
    HORNet & 2.8 & 5.0 & 8.7 & 12.0 & 12.2 \\
    Frame2Clip & 2.7 & 8.2 & 7.8 & 8.0 & 7.8 \\
    \bottomrule
  \end{tabular}
\end{table}

\begin{table}[t]
  \centering\footnotesize
  \setlength{\tabcolsep}{6pt}\renewcommand{\arraystretch}{1.05}
  \caption{Full Video-MME sweep: accuracy (\%) on the standard benchmark (chance $25\%$; majority baseline $27.2\%$). $^{\dagger}$Motion-o, Qwen3-VL-4B, and Gemma-4-12B-it Video-MME are the authors' cited numbers.}
  \label{app:sweep-vmme}
  \begin{tabular}{@{}l ccccc@{}}
    \toprule
    Sampler & $N{=}1$ & $4$ & $8$ & $16$ & $24$ \\
    \midrule
    \multicolumn{6}{@{}l}{\textit{Eagle2.5-8B}}\\
    Uniform & 47.4 & 54.5 & 58.9 & 62.2 & 65.4 \\
    Random & 50.9 & 54.7 & 59.5 & 61.3 & 63.0 \\
    HORNet & 48.6 & 52.8 & 56.9 & 61.6 & 64.2 \\
    \midrule
    \multicolumn{6}{@{}l}{\textit{Qwen3-VL-4B$^{\dagger}$}}\\
    Uniform & 41.6 & 48.6 & 52.8 & 56.6 & 59.7 \\
    Random & 44.0 & 50.9 & 52.6 & 55.3 & 57.8 \\
    HORNet & 43.1 & 46.0 & 49.3 & 53.6 & 55.9 \\
    Frame2Clip & 49.8 & 48.7 & 52.5 & 55.7 & 57.8 \\
    \midrule
    \multicolumn{6}{@{}l}{\textit{Motion-o (7B)$^{\dagger}$}}\\
    Uniform & 41.9 & 48.3 & 51.7 & 54.9 & 56.9 \\
    Random & 44.7 & 48.8 & 50.4 & 54.2 & 56.6 \\
    HORNet & 42.8 & 46.8 & 48.2 & 53.1 & 55.6 \\
    Frame2Clip & 49.9 & 50.1 & 53.1 & 55.8 & 58.2 \\
    \midrule
    \multicolumn{6}{@{}l}{\textit{Gemma-4-12B-it$^{\dagger}$}}\\
    Uniform & 35.9 & 45.4 & 48.6 & 52.1 & 54.1 \\
    Random & 38.9 & 43.5 & 47.6 & 50.7 & 52.1 \\
    HORNet & 35.3 & 42.5 & 46.2 & 51.0 & 53.7 \\
    \midrule
    \multicolumn{6}{@{}l}{\textit{Molmo2-8B}}\\
    Uniform & 47.2 & 52.4 & 56.3 & 59.8 & 60.9 \\
    Random & 49.8 & 53.7 & 56.1 & 60.0 & 60.3 \\
    HORNet & 50.0 & 50.4 & 52.7 & 57.3 & 60.0 \\
    \bottomrule
  \end{tabular}
\end{table}

\end{document}